\documentclass{article}

\usepackage{microtype}
\usepackage{graphicx}
\usepackage{subfigure}
\usepackage{booktabs} 
\usepackage{multirow}

\usepackage{hyperref}

\usepackage[accepted]{icml2023}

\usepackage{amsmath}
\usepackage{amssymb}
\usepackage{mathtools}
\usepackage{amsthm}
\usepackage{xspace}

\usepackage[capitalize,noabbrev]{cleveref}

\theoremstyle{plain}
\newtheorem{theorem}{Theorem}[section]

\theoremstyle{definition}
\newtheorem{definition}[theorem]{Definition}

\theoremstyle{remark}

\newcommand{\datasetFont}{\text}
\newcommand{\ours}{\datasetFont{DP-Flat}\xspace}

\usepackage[textsize=tiny]{todonotes}

\begin{document}

\twocolumn[
\icmltitle{Privacy-preserving Fine-tuning of Large Language Models through Flatness}

\icmlsetsymbol{equal}{*}

\begin{icmlauthorlist}
\icmlauthor{Tiejin Chen}{asu}
\icmlauthor{Longchao Da}{asu}
\icmlauthor{Huixue Zhou}{umn}
\icmlauthor{Pingzhi Li}{ustc}
\icmlauthor{Kaixiong Zhou}{ncsu}
\icmlauthor{Tianlong Chen}{unc}
\icmlauthor{Hua Wei*}{asu}
\end{icmlauthorlist}

\icmlaffiliation{asu}{Arizona State University}
\icmlaffiliation{umn}{University of Minnesota}
\icmlaffiliation{ustc}{University of Science and Technology of China}
\icmlaffiliation{ncsu}{North Carolina State University}
\icmlaffiliation{unc}{University of North Carolina at Chapel Hill}

\icmlcorrespondingauthor{Hua Wei}{hua.wei@asu.edu}


\vskip 0.3in
]

\printAffiliationsAndNotice{}
\begin{abstract}
The privacy concerns associated with the use of Large Language Models (LLMs) have grown recently with the development of LLMs such as ChatGPT. Differential Privacy (DP) techniques are explored in existing work to mitigate their privacy risks at the cost of generalization degradation. Our paper reveals that the flatness of DP-trained models' loss landscape plays an essential role in the trade-off between their privacy and generalization. We further propose a holistic framework to enforce appropriate weight flatness, which substantially improves model generalization with competitive privacy preservation. It innovates from three coarse-to-grained levels, including perturbation-aware min-max optimization on model weights within a layer, flatness-guided sparse prefix-tuning on weights across layers, and weight knowledge distillation between DP \& non-DP weights copies. Comprehensive experiments of both black-box and white-box scenarios are conducted to demonstrate the effectiveness of our proposal in enhancing generalization and maintaining DP characteristics. For instance, on text classification dataset QNLI, \ours achieves similar performance with non-private full fine-tuning but with DP guarantee under privacy budget $\epsilon=3$, and even better performance given higher privacy budgets. Codes are provided in the supplement.
\end{abstract}

\section{Introduction}
Large Language Models (LLMs) such as GPT-4~\cite{OpenAI_GPT4_2023} and Llama 2~\cite{touvron2023llama} have become integral in various real-world applications, including story generation~\cite{zhou2023recurrentgpt,yang2022re3}, AI agents~\cite{mialon2023augmented,da2023open}, chatbots~\cite{luo2022critical} and sim-to-real learning~\cite{da2023llm}. Despite their widespread use, these models raise significant privacy concerns. Previous studies have shown that LLMs can memorize and potentially leak sensitive information from their training data~\cite{carlini2021extracting,mireshghallah2022memorization}, which often includes personal details like emails~\cite{huang2022large}, phone numbers and addresses~\cite{carlini2021extracting}. There are also LLMs trained especially for clinical and medical usage with highly sensitive data~\cite{yang2022gatortron}. The leakage of such information from LLMs may cause a severe privacy issue. 

Differential Privacy (DP) has emerged as a key method for protecting data privacy in LLMs, yet sacrificing the generalization ability. Specifically, techniques such as Differentially Private Stochastic Gradient Descent (DP-SGD)~\cite{abadi2016deep} have been employed to improve the trade-off between privacy and performance. \emph{However, there remains a noticeable performance gap between DP-trained models and standard models}, in both full fine-tuning and parameter-efficient training settings~\cite{li2021large,du2023dp}. Moreover, all current works focus on improving privacy for white-box LLMs, which have limited applicability to closed-source LLMs in real-world scenarios. Therefore, there is an urgent call for pioneering efforts to design effective algorithms in black-box privacy-preserving optimization, which is under-explored in DP-trained LLMs~\cite{malladi2023fine,chen2023instructzero} to our best knowledge.

\begin{figure}[t!]
\centering
\includegraphics[width=0.45\textwidth]{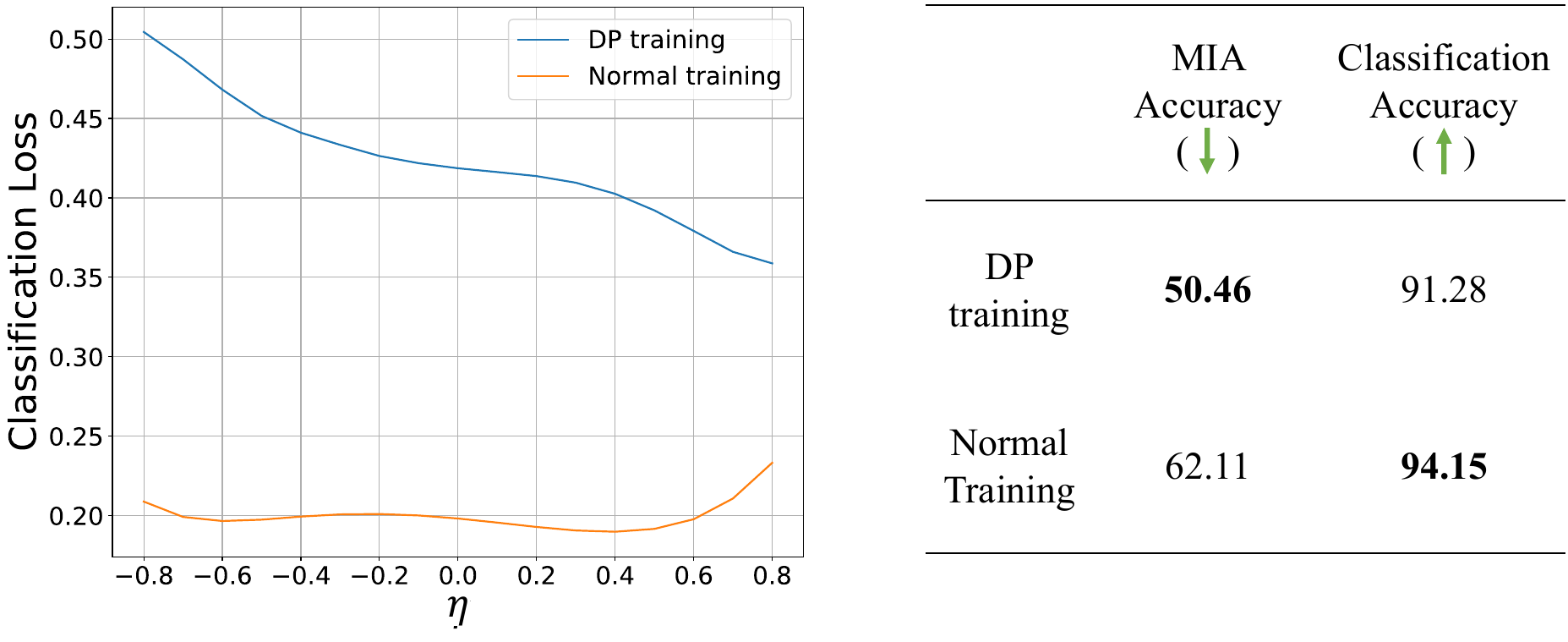}
\caption{Left: Weight loss landscape for DP-trained LLMs and normal (non-private) training on SST-2. The DP-trained model has a sharper loss landscape. Right: The privacy-performance trade-off for DP-trained LLMs: Compared with normal trained models, the DP-trained model has lower privacy risks (better privacy) under Membership Inference Attack (MIA), while it shows lower classification accuracy (worse performance).}
\label{fig:flat}
\end{figure}

\begin{figure*}[t!]
\centering
\includegraphics[width=0.99\textwidth]{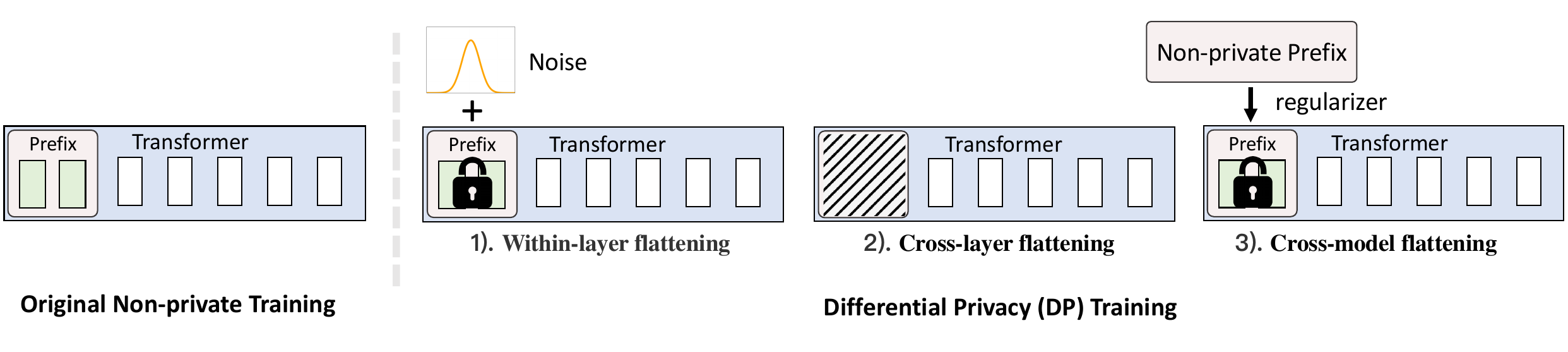}
\vspace{-5mm}
\caption{Our methods improve the flatness of weight loss landscape from three aspects: (1) Within-layer flattening, where a perturbation-aware min-max optimization is utilized to encourage the loss flatness within the weight space of each LLM layer. (2) Cross-layer flattening, where a sparse prefix-tuning algorithm guides layer selection with a flatness-ware indicator. (3) Cross-model flattening, where non-private prefixes are used to guide DP training through knowledge distillation regularization.}
\label{fig:show}
\end{figure*}

To understand this performance gap, we examine the loss landscape of DP-trained models compared to the ones from non-private training. As shown in Figure~\ref{fig:flat}, it illustrates the analysis with the following formula:
$$
f(\eta) = \mathcal{L}(\mathcal{D}\ |\ \mathbf{w} + \eta \cdot \mathbf{d}),
$$
where $\mathcal{D}$ and $\mathbf{w}$ represent the dataset and model weights, respectively, and $\mathbf{d}$ is a random noise sampled from a standard Gaussian distribution and $\eta$ is the magnitude. 
It reveals that DP-trained models tend to have a sharper (\textit{i.e.}, less flatness) loss landscape with respect to model weights. Then, a natural question comes:
\begin{center}
\textit{Q: Does the Loss Flatness Affect the Privacy and Performance Trade-off in DP-trained LLMs?}
\end{center}

If so, could we take one step further --- improving performance with competitive privacy by appropriately enhancing the loss landscape's flatness? We present a holistic framework, consisting of three novel strategies to promote weight-level flatness from three coarse-to-grained perspectives: 

\noindent $\rhd$~\textit{Within-layer flattening}. We introduce a perturbation-aware min-max optimization to encourage the loss flatness within the weight space of each LLM layer.

\noindent $\rhd$~\textit{Cross-layer flattening}. We propose a sparse prefix-tuning algorithm to facilitate the landscape flatness across LLM layers~\cite{li2021prefix}, where a flatness-ware indicator will guide the sparse layer selection.

\noindent $\rhd$~\textit{Cross-model flattening}. We design a novel approach using non-private prefixes to guide DP training through knowledge distillation regularization with non-private weights, aiming to improve the flatness in the whole weight space of LLMs.

Our main contributions can be summarized as follows:

\noindent $\bullet$~We conduct pioneering efforts to investigate the critical role of weight flatness in DP-trained LLMs. We show that appropriately enforced weight flatness improves the performance of DP-trained LLMs with competitive privacy.

\noindent $\bullet$~We propose a holistic framework named \textbf{\ours} to promote weight flatness in three coarse-to-grained levels, including perturbation-ware mix-max optimization on weights within a layer, flatness-guided sparse prefix-tuning on weights across layers, and weight knowledge distillation between DP \& non-private weight copies. 

\noindent $\bullet$ We make pioneering effort on proposing effective privacy-preserving algorithms for closed-source large language models with tailored black-box optimization.

\noindent $\bullet$~Comprehensive experiments in both black-box and white-box settings are conducted to show that our proposed methods can bridge the notorious gap between non-private LLMs and DP-trained LLMs. For example, on text classification dataset QNLI, \ours achieves similar performance with non-private full fine-tuning but with DP guarantee under privacy budget $\epsilon=3$, and it even outperforms non-private full fine-tuning given $\epsilon=8$.

\section{Related Work}
\textbf{Learnable Prompts for LLMs:} Prompt-based learning has gained traction, initially focusing on discrete, task-specific prompts~\cite{shin2020autoprompt}. The shift to continuous, learnable prompts (soft prompts) has led to improved performance~\cite{lester2021power,liu2023gpt}. Unlike traditional prompt tuning, prefix-tuning~\cite{li2021prefix} and P-tuning V2~\cite{liu2021p} incorporate prompts at each transformer layer. For prompt tuning or prefix-tuning, zeroth-order optimization (ZO) methods like ZO-SGD~\cite{spall1992multivariate} are employed for black-box settings without requiring knowing the parameters of the original model. MEZO, introduced by \citet{malladi2023fine}, optimizes ZO-SGD for LLM fine-tuning with lower memory needs. While other works also explored black-box optimization methods for both discrete~\cite{chen2023instructzero} and soft prompts~\cite{sun2022black}, they do not investigate the issue of privacy leakage in the model training. 

\textbf{Privacy Leakage in LLMs:} The potential of Large Language Models (LLMs) to memorize training data poses privacy risks~\cite{mireshghallah2022memorization,carlini2022quantifying,ippolito2022preventing}. Such memorization enables the extraction of private information or even direct reconstruction of training data~\cite{parikh2022canary,huang2022large,carlini2021extracting,zhang2022text,elmahdy2023deconstructing}. Studies have demonstrated the feasibility of recovering keywords or predicting training words from sentence embeddings using auxiliary datasets~\cite{pan2020privacy,song2020information}. A notable advancement was made by \citet{li2023sentence}, who introduced an attack model to enhance the efficacy of attacks on sentence embeddings. Recent comprehensive analyses, including those on GPT-4, further underline the seriousness of this issue~\cite{wang2023decodingtrust,OpenAI_GPT4_2023}. In this paper, we employ the Membership Inference Attack (MIA)~\cite{carlini2022membership,yeom2018privacy,shokri2017membership} to evaluate LLMs' vulnerability to privacy leakage issues. 

\textbf{Differential Privacy for LLMs:} 
Differential privacy (DP)~\cite{dwork2006differential} has emerged as a widely accepted technique for protecting individual privacy
The concept of Differential Privacy (DP) in deep learning was introduced by DP-SGD\cite{abadi2016deep}, with subsequent improvements in DP accounting methods like~\cite{mironov2017renyi,dong2019gaussian}. These methods mainly clip the gradients of each example in a batch and
add random noise to the aggregated gradient. To address the privacy leakage issue in LLMs, multiple techniques~\cite{yu2021differentially,anil2021large,dupuy2022efficient,li2021large,du2023dp,lyu2020differentially,tang2024private} have been proposed in the full finetuning process with DP guarantee. With the full finetuning of LLMs requires significantly more data~\cite{scao2021many}, computational resources~\cite{li2021large} and access to the LLM parameters, recent studies have found that DP prefix-tuning~\cite{duan2023flocks}, DP prompt tuning~\cite{li2023privacy} and other parameter-efficient methods~\cite{yu2021large,bu2022differentially} can match the performance of DP full fine-tuning. In this paper, we provide a systematic view of previous methods with loss landscape.

\section{Methods}
\subsection{Preliminaries and Background}
In this paper, we focus on the approximate-DP~\cite{dwork2014algorithmic} for providing the training process privacy:

\begin{definition}[$(\epsilon,\delta)$-Differential Privacy] 
A algorithm $\mathcal{M}:\mathcal{X} \rightarrow \mathcal{Y}$ is said to be $(\epsilon,\delta)$-DP if for
all adjacent datasets $X,X' \in \mathcal{X}$ and all $Y \in \mathcal{Y}$, we have the guarantee:
$$\mathbb{P}[\mathcal{M}(X) \in Y] \leq e^\epsilon\mathbb{P}[\mathcal{M}(X') \in Y] + \delta,$$
\end{definition}
where $X$ and $X'$ are neighboring datasets if they differ in a single entry, $\epsilon$ is the privacy budget, and $\delta$ is the failure probability. Small values of $\epsilon$ and $\delta$ indicate strong privacy protection. 
For DP training in deep learning, $\mathcal{M}$ refers to any optimization methods and $\mathcal{Y}$ refers to possible parameter space for models. DP algorithm $\mathcal{M}$ aims at ensuring the outputs of two similar datasets are indistinguishable, so that the attacker is not able to know the information about the training dataset given model weights.
The typical DP algorithm could be realized via three interleaved steps: clipping per-sample gradient, sampling a random noise $z \sim N(0,\sigma^2 I)$, and adding $z$ to the accumulated clipped gradient. The variance parameter $\sigma^2$ is determined by several factors including total training steps, $\epsilon$, and $\delta$.



\subsection{Enhancing Flatness in White-box Setting}
It is notorious that DP training often sacrifices a larger degree of model accuracy to gain the required data privacy. In this work, we propose to investigate this trade-off from a novel perspective, i.e., comparing the metric of model flatness before and after DP training. As shown in Figure~\ref{fig:flat}, LLMs under DP training are prone to converge to sharp local minima, where the loss value increases quickly in the neighborhood around model weights. In other words, a slight perturbation in the model weights will lead to poor generalization in unseen data. Many previous work has revealed the strong correlation between sharp local minima and unacceptable accuracy in vision and natural language processing~\cite{chen2021vision, du2021efficient, andriushchenko2022towards}.

To mitigate the negative impact of DP training, we propose a flatness-aware framework, termed as \ours, to enhance the accuracy-privacy trade-off. Specifically, considering a multi-layer white-box model, we smooth the sharp local minima of LLMs comprehensively from three perspectives, including within-layer, cross-layer, and cross-model weight flattening.



\textbf{Within-layer Weight Flattening.} Many pioneering works has been explored to regularize the layer-wise independent weights, among which adversarial weight perturbation (AWP) shows superior results~\cite{wu2020adversarial}. AWP flattens the weight loss landscape and aims to improve adversarial robustness, whereas we adopt it with the intuition that the negative impact of DP noise for model accuracy could be lowered.


Let $\mathbf{w}$ represent the trainable parameters in LLMs, and let $\mathcal{D}$ represent the training dataset. Typically in prefix tuning of LLMs, $\mathbf{w}$ is given by the appending learnable tokens at each layer~\cite{li2021prefix}. AWP updates the model weights with two gradient backpropagation steps:
\begin{equation}
\begin{array} {rll}
    \textbf{v} & = & \arg \max_{\mathbf{v}} \mathcal{L}(\mathcal{D};\mathbf{w}+\mathbf{v}); \\
    \mathbf{w} & \leftarrow & (\mathbf{w}+\mathbf{v}) - \eta \nabla_{\mathbf{w}+\mathbf{v}} \mathcal{L}(\mathcal{D};\mathbf{w}+\mathbf{v}) - \mathbf{v}.
\end{array} 
\end{equation}
The first step seeks perturbation gradient  $\mathbf{v}$ via gradient ascent, which represents the case of worst loss centered around the current weights $\mathbf{w}$. After adversarially applying the perturbation gradient on model (i.e., $\mathbf{w}+\mathbf{v}$), the second step updates the model weights with another complete forward and
backward passes. In this way, the weight loss landscape has a smaller curvature at the
final learned weights, which in turn shrinks the accuracy loss. 

We tailor AWP to DP training with two critical changes. First, we only consider applying the adversarial perturbation gradients in the first $T$ rounds of training, following which the normal model updating is turned on. With this procedure, we can save the external time cost of adversarial computation while guiding model towards smooth loss region. Second, during the initial $T$ rounds, the required noises in DP training are only added to the final gradient $\nabla_{\mathbf{w}+\mathbf{v}} \mathcal{L}(\mathcal{D};\mathbf{w}+\mathbf{v})$, instead of the process of computing $\mathbf{v}$. This ensures the correct location of the adversarial gradient.

\textbf{Cross-layers Weight Flattening.} Beyond the regular weight regularization, we manipulate prefix weights in LLMs to further improve flatness via considering their cross-layer dependencies. In particular, the prefix tuning adds the differential parameters in every layer of LLMs: Given a $n$-layer LLMs, prefix weights $\mathbf{w}_i$ are appended at the $i$-th layer and we have $\mathbf{w} = [\mathbf{w}_1,...,\mathbf{w}_n]$. However, as the prefix added to a layer influences its following output, the flatness of the weight loss landscape is determined by where the prefix modules are added. Thus we explore how to quickly quantify the model sharpness and how to adopt it for controlling the positions of prefix layers.

\begin{definition}[Prefix Sharpness]
\label{def:sharpness}
Given prefix parameters $\mathbf{w}'$ within a box in parameter space $\mathcal{C}_{\eta}$ with sides of length $\eta > 0$,  centered around a minima of interest at parameters $\mathbf{w}$, 
the sharpness of loss  $\nabla \mathcal{L}(\mathbf{w})$ at  $\mathbf{w}$ is defined as: 
\begin{equation*}
\mathrm{Sharpness}:=\frac{\mathrm{max} _{\mathbf{w}' \in \mathcal{C}_{\eta}} \left( \mathcal{L}(\mathbf{w}') - \mathcal{L}(\mathbf{w}) \right)}{1+ \mathcal{L}(\mathbf{w})}.
\label{equ:keskar-sharpness}
\end{equation*}
\end{definition} 
In practice, we approximate the above prefix sharpness  by sampling prefix weights $\mathbf{w}'$: $$\mathbf{w}' \in \{ \mathbf{w} - \eta \nabla \mathcal{L}(\mathbf{w}|\mathcal{D})| \eta \in [0,1]\}.$$ Note that this metric indicates the generalization capability of the included prefixes, where a higher value means the prefix modules can deteriorate the model accuracy after DP training and vice versus. 




Based on the sharpness definition, we design a greedy  solution to gradually eliminate the prefix layers and keep those resulting in lowest sharpness. First, with the prefix initialization at all the layers of LLMs, we can compute the its sharpness value. Next we remove one prefix layer each time and obtain:
\begin{equation}
\label{eq:eli}
    \mathbf{w}_{-i} = [\mathbf{w}_1,..,\mathbf{w}_{i-1},\mathbf{w}_{i+1},..], i = 1\cdots, n.
\end{equation}
For each prefix detaching, we calculate the corresponding sharpness of remaining model parameters. The prefix layer where its removing is associated with the lowest sharpness will be permanently deleted. We will continue this loop until the remaining prefixes meet our sparse requirement or the sharpness metric does not decrease. We get all the sharpness results right after the same random initialization and do not require fine-tuning. After this greedy procedure, LLMs are appended with the sparse prefixes only at the chosen layers and used for DP training. 


\textbf{Cross-models Weight Flattening.} Recall that DP training inevitably results in a sharper loss landscape than that of normal training. One of the intuitive ways to generalize DP-guaranteed model is to regularize it with the normal counterpart via knowledge distillation~\cite{ gou2021knowledge}. 
For this purpose, given parameters $\textbf{w}$ fine-tuned with DP framework, we create their duplicates $\textbf{w}_{\mathrm{nor}}$ using the same network architecture and initialization but fine-tuning them normally. We then define a new term of loss function to force the weight closeness between $\textbf{w}$ and $\textbf{w}_{\mathrm{nor}}$:
\begin{equation}
    \mathcal{L}_g = \lVert \textbf{w} - \textbf{w}_{\mathrm{nor}} \rVert_2.
\end{equation}
Therefore, the final loss function will be:
\begin{equation}
\label{eq:final}
    \mathcal{L}_f = \mathcal{L}(\mathcal{D}|\mathbf{w})+ \lambda \mathcal{L}_g,
\end{equation}
where $\mathcal{L}$ can be any loss function in general, such as cross-entropy loss for sentence classification tasks, and $\lambda$ is the balancing factor for regularization. It is minimized using DP training to achieve both data privacy protection and the desired accuracy. Finally, we summarize our training pipeline for white-box setting in \cref{alg:white}.

\begin{algorithm}[h!]
   \caption{\ours on White-box training pipeline}
   \label{alg:white}
   \small
\begin{algorithmic}[1]
   \STATE {\bfseries Input:} $\lambda$, $\eta$,warm-up epochs $E$, DP training total epochs $T_{dp}$, normal training epochs $T_{nor}$, elimination rounds $R$, random initialization prefix $\mathbf{w}= [\mathbf{w}_1,...,\mathbf{w}_n]$.
   \IF{\textit{Cross-layers Weight Flattening}}
   \FOR{$r=1$ {\bfseries to} $R$}
   \STATE $S_{\min} = \infty$
   \STATE $P = 0$
   \FOR{$i=1$  {\bfseries to} $n$}
   \STATE Get  $\mathbf{w}_{-i}$ in \cref{eq:eli}
   \STATE Compute sharpness $S$ for $\mathbf{w}_{-i}$
   \IF{$S < S_{\min}$}
   \STATE  $S_{\min} = S$, $P=i$
   \ENDIF
   \ENDFOR
   \STATE  $\mathbf{w} \leftarrow  \mathbf{w}_{-P} $
   \ENDFOR
   \ENDIF
   \STATE $\mathbf{w}_{nor} = \mathbf{w}$
   \FOR{$t=1$ {\bfseries to} $T_{\mathrm{nor}}$}
   \STATE $\mathbf{w}_{\mathrm{nor}} \leftarrow \mathbf{w}_{\mathrm{nor}}-\eta \nabla_{\mathbf{w}_{\mathrm{nor}}} \mathcal{L}(\mathcal{D}|\mathbf{w}_{\mathrm{nor}}) $
   \ENDFOR
   \FOR{$t=1$ {\bfseries to} $T$}
   \IF{$t <= E$ \AND \textit{Within-layer Weight Flattening}}
   \STATE Compute $\mathbf{v}$
   \STATE  $\mathcal{L}_f = \mathcal{L}
   (\mathcal{D}|\mathbf{w}+\mathbf{v})$
   \ELSE
   \STATE $\mathcal{L}_f = \mathcal{L}(\mathcal{D}|\mathbf{w})$
   \ENDIF
   \IF{\textit{Cross-model Weight Flattening}}
   \STATE $\mathcal{L}_f = \mathcal{L}_f + \lambda  \lVert \textbf{w} - \textbf{w}_{\mathrm{nor}} \rVert_2$
   \ELSE
   \STATE $\mathcal{L}_f = \mathcal{L}_f$
   \ENDIF
   \STATE Update $\mathbf{w}$ with $\mathcal{L}_f$ and DP-Adam
   \ENDFOR
\end{algorithmic}
\end{algorithm}

\begin{figure}[t!]
\centering
\vspace{-3mm}
\includegraphics[width=0.45\textwidth]{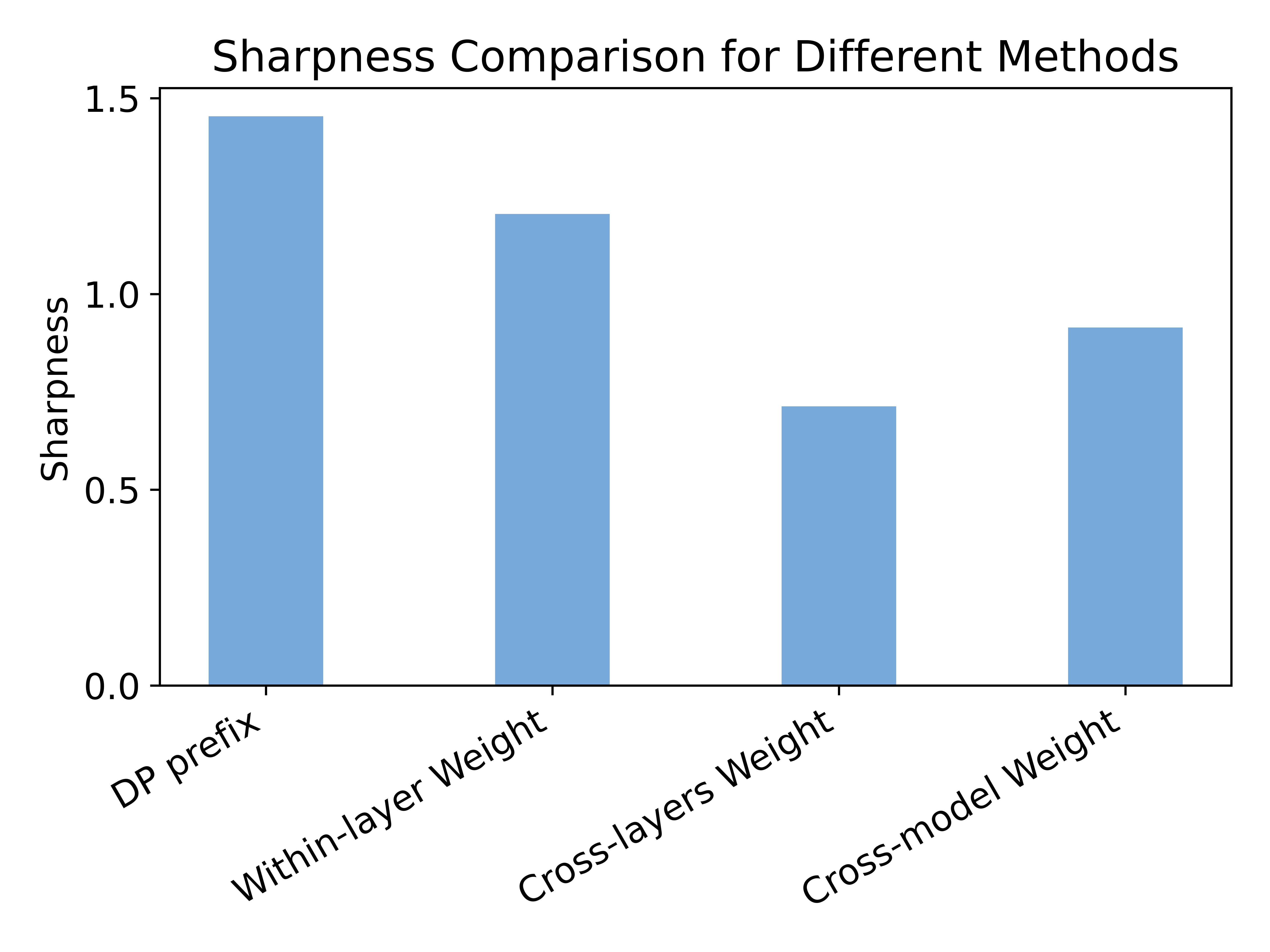}
\caption{Sharpness for DP trained prefix tuning plus our proposed three weight flattening methods on SST-2. Our proposed model has a flatter loss landscape.}
\label{fig:three-flat}
\end{figure}

\subsection{Analysis of Sharpness over Landscape}
We calculate the sharpness over the landscape for proposed methods on SST-2 by integrating the proposed weight flattening method with DP-trained prefix tuning. The results in \cref{fig:three-flat} show that all our proposed three weight flattening methods flatten the weight loss landscape. This matches the design intuition that \ours smooths the sharp local minima of LLMs comprehensively by perturbation on weights within a layer, flatness-guided sparse prefix tuning across layers, and weight flattening with regularization. Later in~\cref{sec:exp}, we will validate how \ours improves the performance with competitive privacy by enhancing the loss landscape. 

\subsection{Enhancing Flatness in Black-box Setting}
While LLMs of interest are oftentimes black boxes, i.e., their weights are not accessible for training,  
in this section, 
we extend our \ours framework to the black-box settings.

To deal with black-box settings of neural networks, the zeroth-order~(ZO) optimizers~\cite{malladi2023fine} are often used to estimate the gradient of neural networks using finite differences without any backpropagation. To enable DP guarantee for black-box LLMs, DPZero~\cite{zhang2023dpzero} were proposed: Let $\mathbf{g}$ represent the noise sampled from Gaussian distribution $\mathcal{N}(0,\sigma^2)$, the gradient will be updated through the following equation: 
\begin{equation}
\label{eq:dpzo}
    \hat\nabla\mathcal L(\mathbf \theta;\mathcal B) =  (\frac{\mathcal L(\mathbf \theta + \varepsilon\mathbf z;\mathcal B) - \mathcal L(\mathbf \theta - \varepsilon\mathbf z;\mathcal B)}{2\varepsilon}+\mathbf{g})\mathbf z.
\end{equation}
Here, $\mathbf{z}$ is a random noise sampled from standard Gaussian distribution, $\varepsilon$ is the perturbation scale and $\mathcal{B}$ represents the batch data. 
DPZero only adds noises to one dimension while directly combining DP with zeroth-order optimization adds noises to the dimensions same with the gradient, which is hundreds of dimensions. \citet{zhang2023dpzero} proves that the performance suffers from the noises added to multi-dimensions. \citet{zhang2023dpzero} also proves with the same variance $\sigma^2$, DPZero achieves the same DP guarantee compared with DP-SGD.

In the black-box setting, we consider improving through non-DP duplication. Compared with the white-box setting, $\textbf{w}_{nor}$ is also trained with the black-box setting, and we treat \cref{eq:dpzo} as the former part in \cref{eq:final} to adopt our methods.  Note that in the black-box setting, it is impractical to improve the across layer weight flatness since we do not have access to the internal weights of each layer in LLMs. It is also difficult to enhance the within-layer weight flatness since the min-max training framework with zeroth order optimization suffers from the high variance of an additional gradient estimation to compute the $\mathbf{v}$~\cite{zhang2022robustify}. Though ablating these two components, we empirically found in experiments that our \ours still delivers the outperforming accuracy under the same privacy constraint.

\section{Experiments}
\label{sec:exp}
\subsection{Experimental Settings}


\paragraph{Datasets} 
To assess the effectiveness of our proposed model, \ours, we explore two principal NLP tasks, i.e., text classification and text generation, across 7 datasets: 
(1) For \emph{text classification}, we engage with datasets from the GLUE benchmark~\cite{wang2018glue}:
SST-2~\cite{socher2013recursive} for sentiment classification;
MNLI~\cite{williams2017broad} and QNLI~\cite{wang2018glue} for sentence pair classification; QQP and TREC~\cite{voorhees1999trec} for topic classification. (2) For \emph{text generation}, we utilize E2E~\cite{novikova2017e2e} and DART~\cite{nan2020dart} for table-to-text generation. 
This selection of datasets allows us to comprehensively evaluate \ours across a spectrum of linguistic tasks and complexities.

\begin{figure*} [h!]
\centering
\begin{tabular}{cccc}
\includegraphics[width=0.32\textwidth]{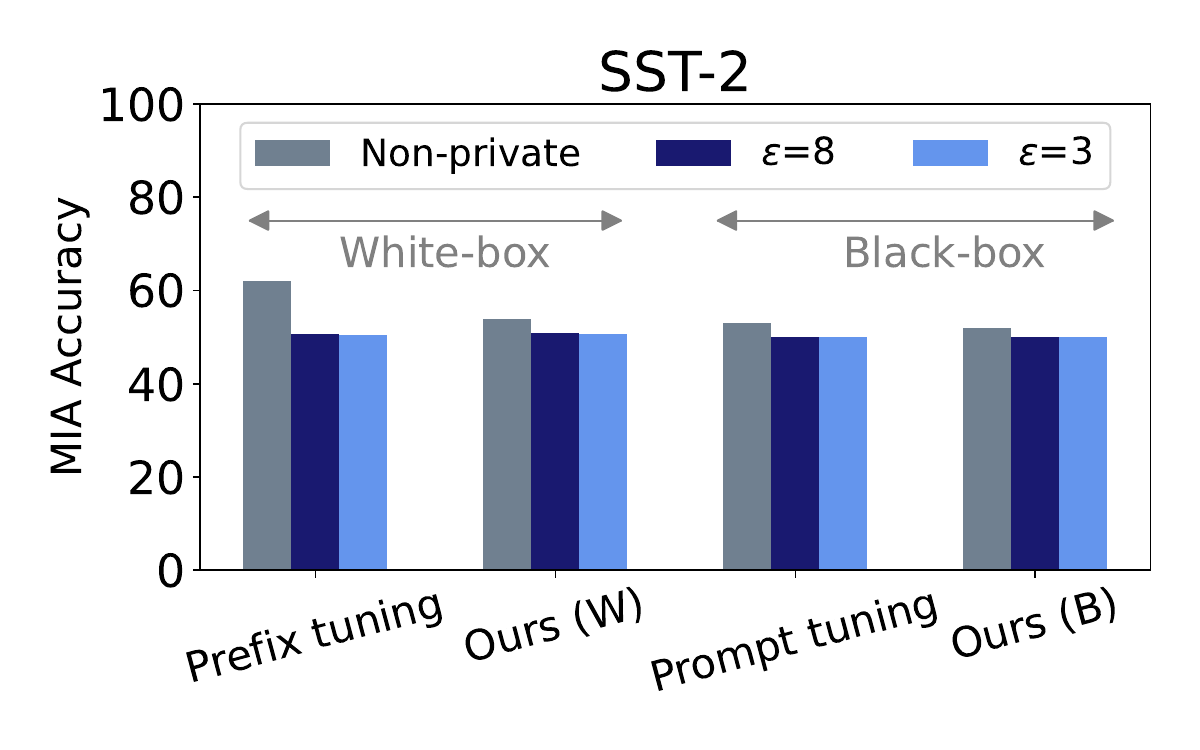} &
\includegraphics[width=0.32\textwidth]{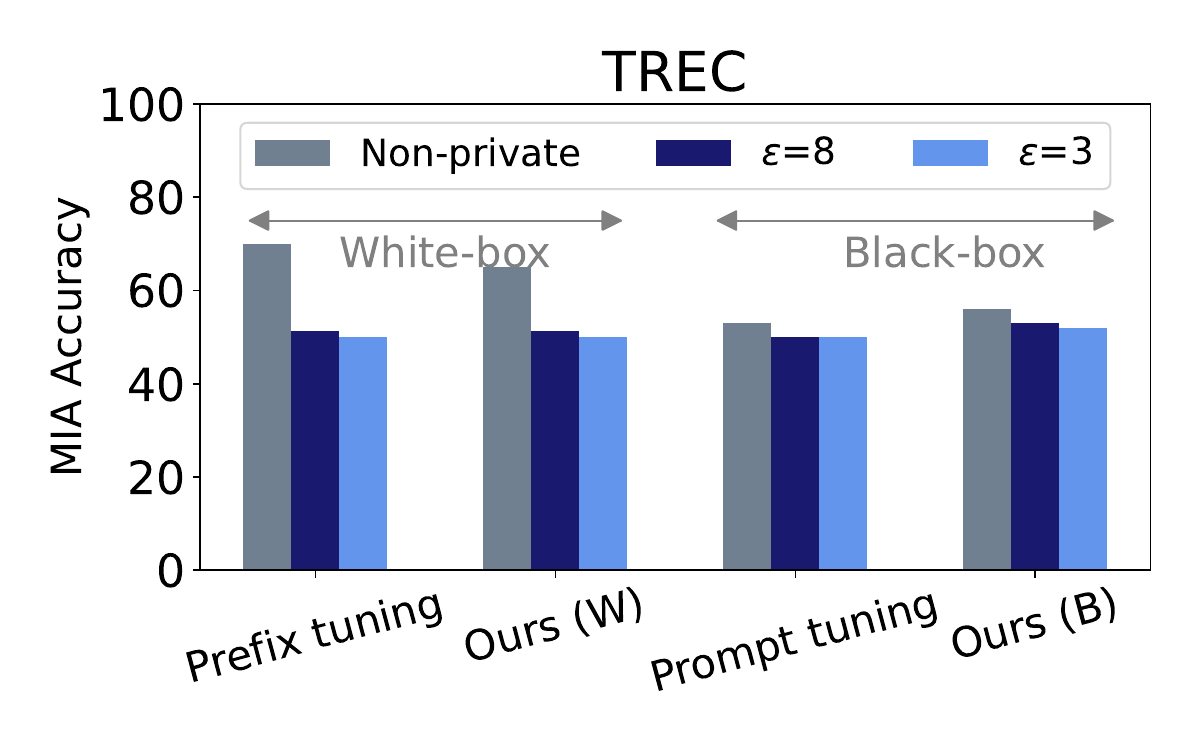} &
\includegraphics[width=0.32\textwidth]{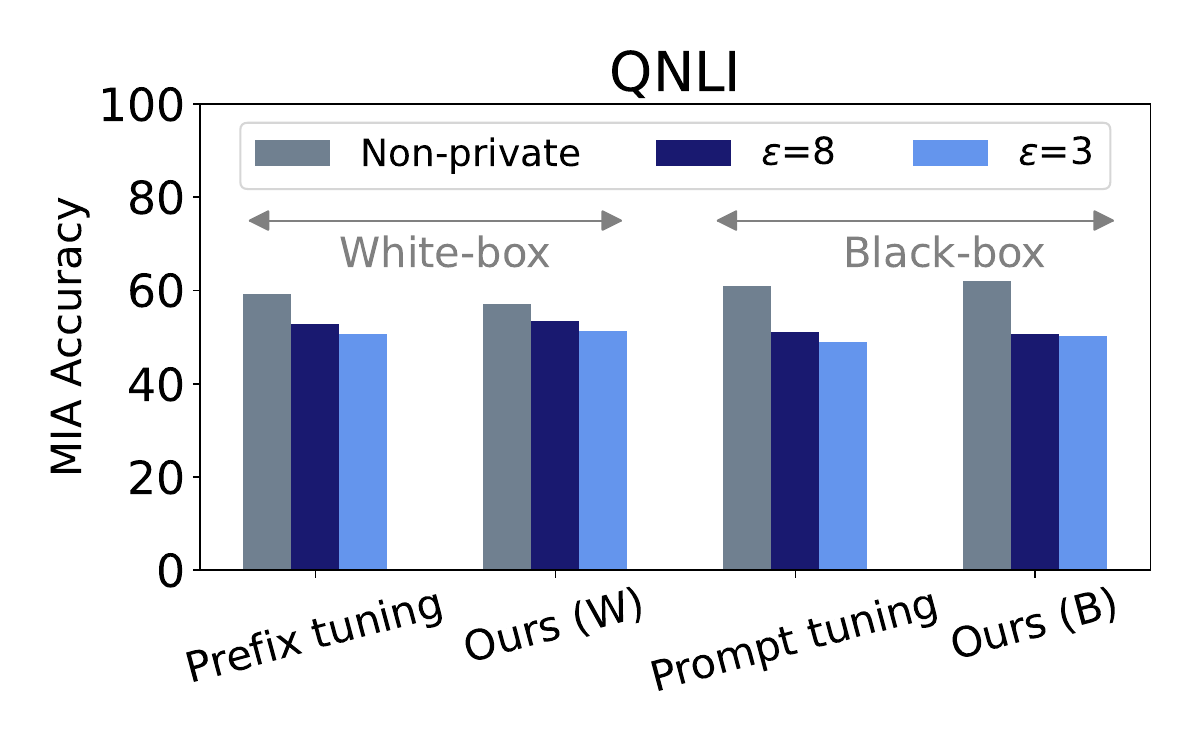} \\[6pt]
\end{tabular}
\vspace{-5mm}
\caption{Comparison of MIA accuracy under both white-box and black-box settings across text classification datasets. The lower the accuracy, the lower the privacy risk. The results show that our proposed method will not affect the privacy protection for both white-box and black-box settings.}
\label{tab:MIA}
\end{figure*}

\textbf{Setups} In the white-box setting, we mainly use Roberta-base~\cite{liu2019roberta} and BERT~\cite{devlin2018bert} for encoder-only architectures and GPT-2~\cite{radford2019language} for decoder-only architectures. In the black-box setting, we adopt Roberta-base. 
For the DP algorithms, we follow the common practice to set the privacy budget as $\epsilon = [3,8]$ and $\delta = \frac{1}{2|\mathcal{D}|}$ for all settings, and we account for privacy through Renyi differential privacy~\cite{mironov2017renyi}. For \ours, we set the regularization weight $\lambda$ in \cref{eq:final} to 0.01 for all experiments. More experiment details are
deferred to Appendix~\ref{app:hype}.

\paragraph{Baselines}
For white-box settings, we mainly compare with full fine-tuning and prefix tuning under non-private and DP training; for black-box settings, we compare with prompt-tuning under non-private and DP training. For the non-private training, it could be considered as DP training with privacy budget $\epsilon=\infty$.

\subsection{Empirical Evaluation of Privacy Risks}
In this section, we conduct experiments to show that \ours shows a similar capability in privacy-preserving as vanilla DP training. Following existing work~\cite{yu2021large, dupuy2022efficient}, we evaluate the privacy risks empirically by membership
inference attack (MIA), 
which targets judging whether a data sample belongs to a training set or not. In this paper, we consider a simple but efficient loss-based MIA~\cite{yeom2018privacy}, which considers the samples with a loss lower than a threshold as the training dataset. A model with higher accuracy in MIA indicates higher privacy risks since the successes mean that the attackers may be able to reveal information about the data used to train the model. More detailed descriptions on the MIA setting can be found in the Appendix~\ref{app:mia}. 
From the results in~\cref{tab:MIA}, we have the following observations: 



(1) Compared with non-private training ($\epsilon=\infty$), all DP baselines show lower accuracies against MIA, indicating better protection. This matches the conclusions in existing DP literature that DP training lowers privacy risks~\cite{zhang2023dpzero,du2023dp,li2023privacy}.

(2) Under the same privacy budget, \ours shows very similar MIA accuracies with DP-trained prefixes and DP full fine-tuning, indicating that \ours does not hurt the privacy protection and the DP guarantee. This is because our proposed methods are designed with DP guarantees while the weight flattening mechanism helps with the generalization capability on classification or generation tasks. 

\vskip -0.1in


\begin{table*}[th]
\centering
\begin{tabular}{cccccccccccc}
\toprule
\multirow{2}{*}{Method}        & \multicolumn{5}{c}{Roberta-base}    &  & \multicolumn{5}{c}{BERT}              \\
                     \cmidrule{2-6} \cmidrule{8-12}
              & MNLI  & QNLI  & SST-2 & QQP   & TREC  && MNLI  & QNLI  & SST-2 & QQP   & TREC  \\ \midrule
 &\multicolumn{11}{c}{Non-private ($\epsilon=\infty$)}                                                                     \\ \midrule
Full Fine-tuning     & 85.95 & 91.06 & \textbf{94.68} & \textbf{88.05} & 93.00 && \textbf{83.09} & \textbf{88.94} & \textbf{91.85} & \textbf{90.17} & 92.60 \\
Prefix Tuning        & \textbf{86.12} & \textbf{91.59} & 94.15 & 87.79 & 91.40 && 79.95 & 86.34 & 91.62 & 89.25 & \textbf{96.00} \\ \midrule
 &\multicolumn{11}{c}{ $\epsilon = 3$}                                                                         \\ \midrule
Full Fine-tuning     & 80.95 & 86.03 & 92.08 & 83.61 & 79.00 && \textbf{72.57} & \textbf{81.70} & 87.50 & \textbf{81.46} & \textbf{73.60} \\
Prefix Tuning        & 79.03 & 83.70 & 91.28 & 80.13 & 78.40 && 60.07 & 65.15 & 81.19 & 71.99 & 48.40 \\
\ours & \textbf{84.12} & \textbf{90.72} & \textbf{93.57} & \textbf{86.05} & \textbf{82.20} &&   65.32    & 71.02 & \textbf{88.53} &   74.68    &   47.80   \\ \midrule
 &\multicolumn{11}{c}{$\epsilon = 8$}                                                                         \\ \midrule
Full Fine-tuning     & 81.42 & 86.03 & 92.18 & 83.61 & 85.40 && \textbf{73.64} & \textbf{82.37} & 88.30 & \textbf{81.92} & \textbf{80.60} \\
Prefix Tuning        & 79.56 & 84.64 & 91.51 & 81.02 & 86.80 && 62.72 & 67.62 & 82.34 & 72.46 &  61.80     \\ 
\ours & \textbf{85.30} & \textbf{91.29} & \textbf{94.03} & \textbf{87.13} & \textbf{90.60} && 67.42     & 72.08 & \textbf{89.56} &   74.29    &   70.20   \\
\bottomrule
\end{tabular}
\caption{Performance of our weight flattening methods with baselines for the sentence classification task w.r.t accuracy on white-box settings across different language models. The higher, the better. The \textbf{best} performance under the same DP training is highlighted. The results show that \ours can increase the performance of DP-trained LLMs for various text classification tasks.}
\label{tab:classification_white_rob}
\end{table*}

\subsection{Evaluation in Classification and Generation}
There is a range of DP methods for LLMs. Does \ours lead to better results under various classification and generation tasks? We conduct experiments under both black-box and white-box settings. We report the test accuracy in classification tasks, and the BLEU score and ROUGE-L for generation tasks.

\begin{table}[h!]
\centering
\small
\resizebox{0.48\textwidth}{!}{
\begin{tabular}{cccccc}
\toprule
\multirow{2}{*}{Method} &  \multicolumn{2}{c}{E2E} && \multicolumn{2}{c}{DART} \\ 
\cmidrule{2-3} \cmidrule{5-6}
                           & BLEU      & ROUGE-L     && BLEU       & ROUGE-L     \\ 
                           \midrule
                         &\multicolumn{5}{c}{Non-private ($\epsilon=\infty$)}  
                         \\ \midrule
Full Fine-tuning          & \textbf{66.59}     & \textbf{69.54}       && \textbf{43.16}      & \textbf{57.85}       \\
Prefix Tuning             & 64.79     & 68.24       && 37.08      & 53.35       \\ \midrule
                         &\multicolumn{5}{c}{$\epsilon=3$}  
                         \\ \midrule
Full Fine-tuning          & 60.3      & 65.31       && 30.75      & 51.69       \\
Prefix Tuning             & 58.2      & 64.51       && 30.26      & 51.43       \\ 
\ours     & \textbf{62.13}     & \textbf{65.84}      & & \textbf{33.14}      & \textbf{52.40}       \\ \midrule
                         &\multicolumn{5}{c}{$\epsilon=8$}  
                         \\ \midrule
Full Fine-tuning          & 62.9      & 66.69       && 32.92      & 53.43       \\ 
Prefix Tuning             & 62.7      & 67.19       && 33.45      & 53.45       \\ 
\ours       & \textbf{64.30}     & \textbf{67.22}       && \textbf{37.06}      & \textbf{53.49}        \\ \bottomrule
\end{tabular}
}
\caption{Comparison of our weight smooth methods with baselines for the table-to-text task on GPT2 and white-box settings. The higher, the better. The \textbf{best} performance under the same DP training is highlighted. \ours performs consistently better than other DP-trained methods on various text generation tasks.}
\label{tab:generation_white}
\end{table}

\subsubsection{White-box Setting}

\paragraph{Text Classification} We first explore whether \ours can bridge the gap between DP-trained models and non-private models ($\epsilon=\infty$) in a white-box setting. In \cref{tab:classification_white_rob}, we provide the results of the experiment for Roberta-base with different tasks. We have the following observations:

(1) \ours can increase the performance of DP prefix tuning significantly, and even outperforms full fine-tuning. Compared with DP prefix tuning, \ours improves at most $8.39\%$ on QNLI and at least $2.5\%$ on SST-2. Furthermore \ours can even beat DP full fine-tuning in all settings with the same trainable parameter with DP prefix tuning. Though \ours is not the best performance on BERT, \ours still shows an improvement over prefix tuning, enjoying a much lower memory cost than full fine-tuning. This is because \ours considers three flattening aspects that can mitigate the negative impact of DP training and achieve a better trade-off between privacy and performance.

(3) \ours can also work well across different models. We train \ours on Bert-base with the same tasks for Roberta-base. Similar performances as in Roberta-base are also found in BERT: \ours outperform DP prefix tuning and DP full tine-tuning in all settings and bridge the gap between non-private models ($\epsilon=\infty$) and DP-trained models.

(2) Though prefix tuning can outperform full fine-tuning in some tasks under Roberta, there is no consistent winner between DP prefix tuning and DP full fine-tuning considering their performance on all datasets, which is coherent with the conclusion made in the previous work~\cite{li2021large}. In comparison, \ours achieves consistently the best performance under Roberta-base. 

\paragraph{Text Generation} For the table-to-text generation, where LLMs are asked to generate the natural language description for the given table entry. We adopt the decoder-only GPT2 for this task and the results are shown in \cref{tab:generation_white}. We have the following observations:

(1) \ours outperforms DP-trained models across all datasets in private training settings. With the same privacy budget $\epsilon$, \ours consistently performs the best. 

(2) The performance of DP training models increases higher privacy budget $\epsilon$, while \ours achieve competitive performance with DP prefix tuning methods with higher $\epsilon$. This indicates that \ours can provide a strong utility for conservative privacy budgets. 

(3) For tasks with different difficulties, \ours shows competitive or better performances. In simple tasks (E2E dataset), the gap between DP-trained models and non-private models ($\epsilon=\infty$) is small. When $\epsilon = 8$, \ours can even compete with prefix tuning with non-private training. For difficult tasks (DART dataset), the performance gap between the non-private model and the DP-trained model becomes much larger. The performance of DP prefix tuning can compete or become even better than DP full fine-tuning, indicating the advantages of full fine-tuning rely on the easy dataset.

\begin{table}[t!]
\small
\centering
\resizebox{0.48\textwidth}{!}{
\begin{tabular}{ccccc}
\toprule
\multirow{2}{*}{Method} & \multicolumn{4}{c}{Roberta-base}  \\
\cmidrule{2-5}
 & MNLI & QQP & SST-2 & TREC  \\
\midrule
& \multicolumn{4}{c}{Non-private ($\epsilon=\infty$)} \\
\midrule
Prompt Tuning with MEZO & 64.51 & 60.93 & 88.46 & 70.61  \\
\midrule
& \multicolumn{4}{c}{$\epsilon=3$} \\
\midrule
Prompt Tuning with DPZero & 53.99 & \textbf{53.41} & 85.2 & 52.14  \\
\ours & \textbf{55.07} & 53.22 & \textbf{86.12} & \textbf{55.46} \\
\midrule
& \multicolumn{4}{c}{$\epsilon=8$} \\
\midrule
Prompt Tuning with DPZero & 55.41& \textbf{53.51} & 86.35 &  53.02\\
\ours & \textbf{57.13} & 53.42 & \textbf{87.38} & \textbf{56.44}\\
\bottomrule
\end{tabular}
}
\caption{Comparison of our flattening methods with baselines for the sentence classification task on black-box setting. The higher, the better. The \textbf{best} performance under the same DP training is highlighted. Under the black-box setting, only prompt tuning could be implemented. \ours achieves competitive performance under different text classification tasks.}
\label{tab:classification_black}
\end{table}

\subsubsection{Black-box Setting} 
In this section, we test \ours in the black-box setting where we can only manipulate input embedding. Therefore, instead of prefix tuning, only prompt tuning could be implemented. 
We compare with the following baseline methods for prompt tuning: (1) non-private prompt tuning with zeroth order optimization method MEZO~\cite{malladi2023fine}, (2) DP prompt tuning with zeroth order optimization method DPZero~\cite{zhang2023dpzero}.
The results are shown in \cref{tab:classification_black} across different datasets with Roberta-base.  We have the following observations:

(1) Compared with the white-box setting, all classification performance decreased under the black-box setting, indicating that zeroth-order optimization for black-box setting remains a challenging problem. The bridge between the non-private model and the DP-trained model in the black-box setting becomes bigger than the bridge white-box setting in general, indicating further effort should be made to improve the stability in the black-box setting. 

(2) \ours remains comparable performance for the QQP dataset and consistently improves the performance in all other datasets. Despite the difficulties of black-box settings in calculating the gradients, \ours still shows better accuracy under the same privacy constraint. This is because \ours enhances the flatness through regularizations with non-private duplications.

\subsection{Detailed Studies}
In this section, we conduct detailed studies on how different flatness aspects and different $\lambda$ in \cref{eq:final} influence the final performance.

\begin{figure}[t!]
\centering
\includegraphics[width=0.38\textwidth]{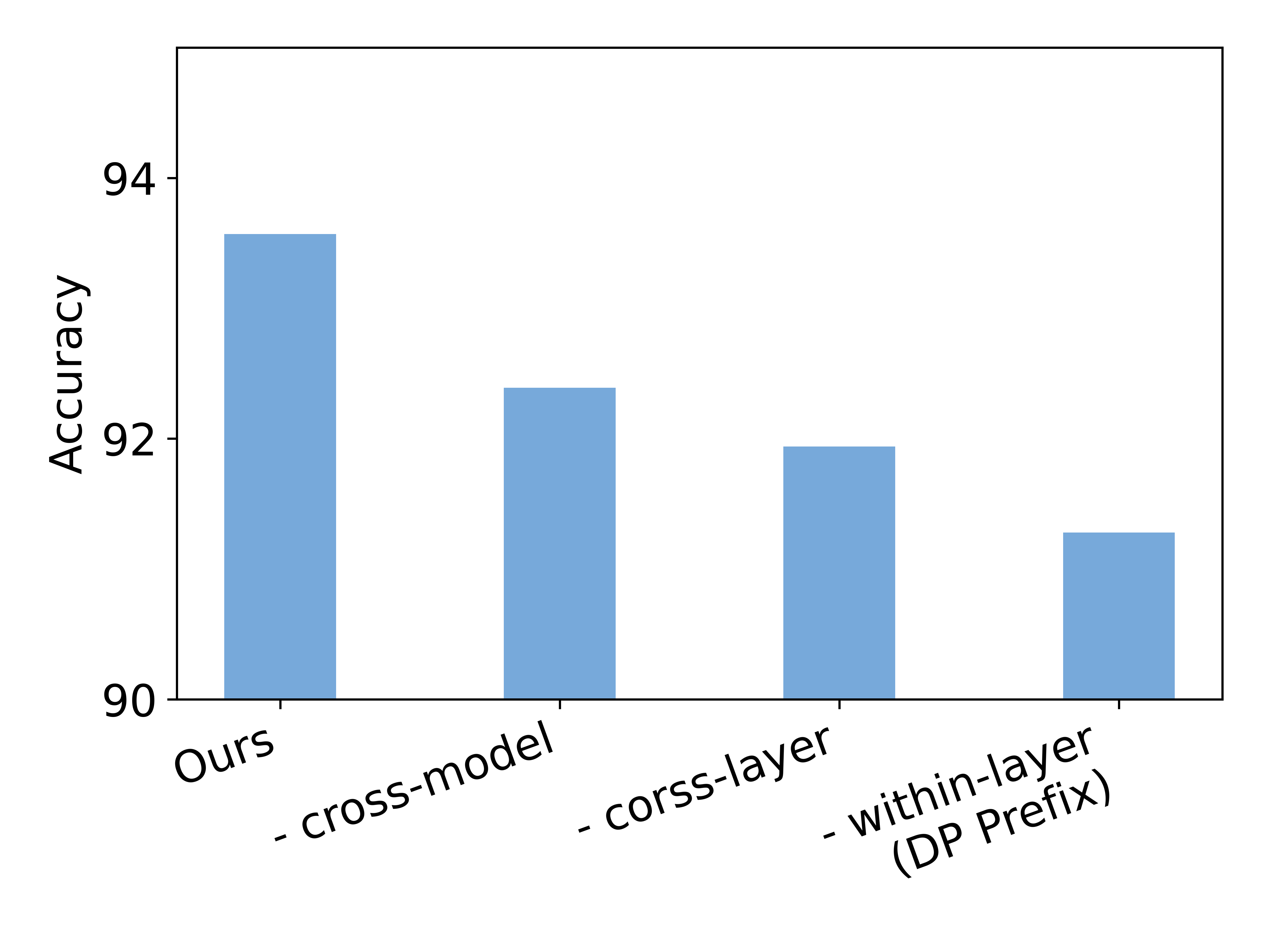}
\caption{Influences of gradually removing different flatness methods on the classification performance w.r.t. accuracy under SST-2 dataset on Roberta-base. The higher, the better. Each aspect helps the final performance while maintaining the DP guarantee.}
\label{fig:ablation}
\end{figure}

\paragraph{Ablation Study on Different Flatness Aspects} To test how much each part of \ours contributes to the final results, we conduct ablation studies to show the performance while gradually removing our methods. Specifically, we conduct experiments on QNLI and SST-2 on Roberta-base. 
\cref{fig:ablation} shows the performance of variants of our
method. We can see that each component will help the performance while maintaining the privacy guarantee, indicating the effectiveness of the proposed flattening methods. Note that our method will downgrade to DP-trained prefix tuning when all three aspects are removed.

\begin{figure}[t!]
\centering
\includegraphics[width=0.35\textwidth]{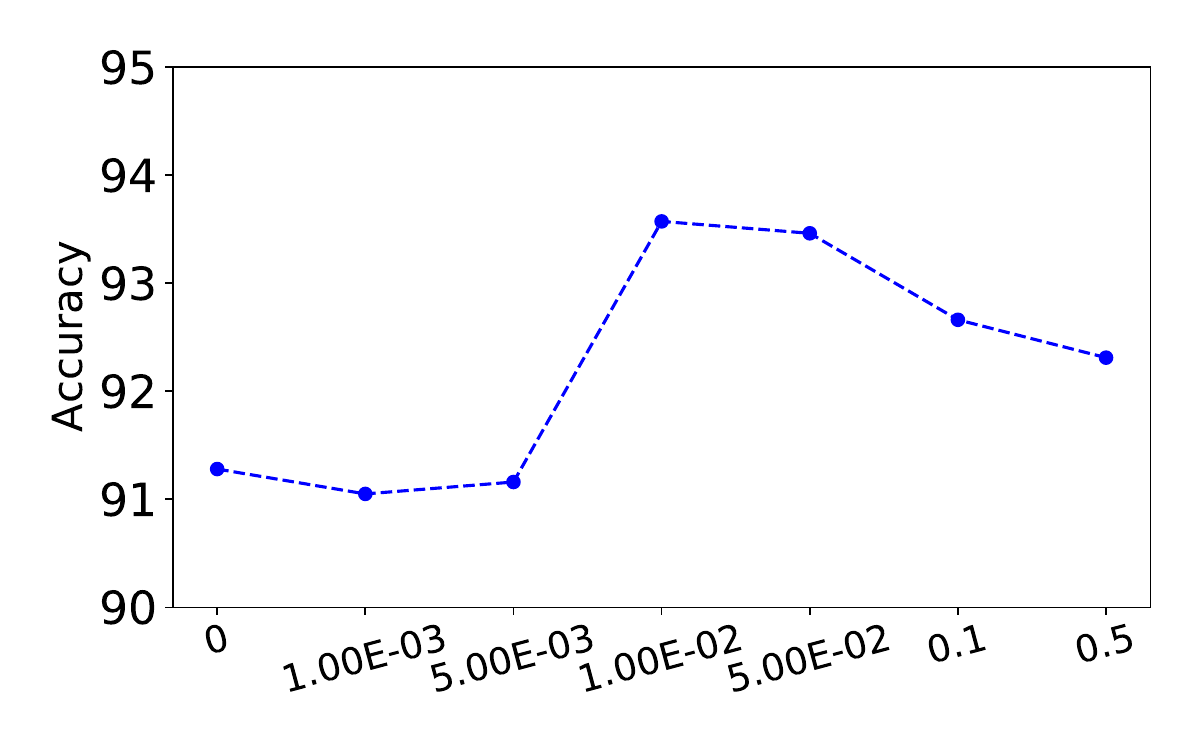}
\caption{Influences of different values of factor $\lambda$ on the classification performance w.r.t. accuracy under SST-2 dataset on Roberta-base. The higher, the better.}
\label{fig:sensitivity}
\vskip -0.2in
\end{figure}

\paragraph{Sensitivity on Different $\lambda$} The regularization factor in \cref{eq:final} balances the flattening with knowledge distillation and DP training. As is shown in~\cref{fig:sensitivity},
when we use knowledge distillation, \ours performs better \ours without knowledge distillation. Note that when $\lambda=0$, our method will not consider cross-model flattening. In this paper, we set $\lambda$ as $1e^{-2}$ as it performs the best empirically.

\section{Conlusion}

In this paper, we address the challenge of balancing data privacy with performance in Large Language Models (LLMs) through Differential Privacy (DP). We introduce a novel framework aimed at enhancing the flatness of the loss landscape in DP-trained models, proposing strategies at three levels: within-layer flattening, cross-layer flattening, and cross-model flattening. Our approach effectively narrows the performance gap between DP-trained LLMs and their standard counterparts, offering pioneering solutions for privacy-preserving algorithms in closed-source settings. Our comprehensive experiments demonstrate significant performance improvements across different tasks in both black-box and white-box settings, highlighting the feasibility and effectiveness of our methods in enhancing the utility of DP-trained LLMs for real-world applications. 


\nocite{langley00}

\bibliography{example_paper}

\begin{thebibliography}{66}
\providecommand{\natexlab}[1]{#1}
\providecommand{\url}[1]{\texttt{#1}}
\expandafter\ifx\csname urlstyle\endcsname\relax
  \providecommand{\doi}[1]{doi: #1}\else
  \providecommand{\doi}{doi: \begingroup \urlstyle{rm}\Url}\fi

\bibitem[Abadi et~al.(2016)Abadi, Chu, Goodfellow, McMahan, Mironov, Talwar,
  and Zhang]{abadi2016deep}
Abadi, M., Chu, A., Goodfellow, I., McMahan, H.~B., Mironov, I., Talwar, K.,
  and Zhang, L.
\newblock Deep learning with differential privacy.
\newblock In \emph{Proceedings of the 2016 ACM SIGSAC conference on computer
  and communications security}, pp.\  308--318, 2016.

\bibitem[Andriushchenko \& Flammarion(2022)Andriushchenko and
  Flammarion]{andriushchenko2022towards}
Andriushchenko, M. and Flammarion, N.
\newblock Towards understanding sharpness-aware minimization.
\newblock In \emph{International Conference on Machine Learning}, pp.\
  639--668. PMLR, 2022.

\bibitem[Anil et~al.(2021)Anil, Ghazi, Gupta, Kumar, and
  Manurangsi]{anil2021large}
Anil, R., Ghazi, B., Gupta, V., Kumar, R., and Manurangsi, P.
\newblock Large-scale differentially private bert.
\newblock \emph{arXiv preprint arXiv:2108.01624}, 2021.

\bibitem[Bu et~al.(2022)Bu, Wang, Zha, and Karypis]{bu2022differentially}
Bu, Z., Wang, Y.-X., Zha, S., and Karypis, G.
\newblock Differentially private bias-term only fine-tuning of foundation
  models.
\newblock \emph{arXiv preprint arXiv:2210.00036}, 2022.

\bibitem[Carlini et~al.(2021)Carlini, Tramer, Wallace, Jagielski, Herbert-Voss,
  Lee, Roberts, Brown, Song, Erlingsson, et~al.]{carlini2021extracting}
Carlini, N., Tramer, F., Wallace, E., Jagielski, M., Herbert-Voss, A., Lee, K.,
  Roberts, A., Brown, T., Song, D., Erlingsson, U., et~al.
\newblock Extracting training data from large language models.
\newblock In \emph{30th USENIX Security Symposium (USENIX Security 21)}, pp.\
  2633--2650, 2021.

\bibitem[Carlini et~al.(2022{\natexlab{a}})Carlini, Chien, Nasr, Song, Terzis,
  and Tramer]{carlini2022membership}
Carlini, N., Chien, S., Nasr, M., Song, S., Terzis, A., and Tramer, F.
\newblock Membership inference attacks from first principles.
\newblock In \emph{2022 IEEE Symposium on Security and Privacy (SP)}, pp.\
  1897--1914. IEEE, 2022{\natexlab{a}}.

\bibitem[Carlini et~al.(2022{\natexlab{b}})Carlini, Ippolito, Jagielski, Lee,
  Tramer, and Zhang]{carlini2022quantifying}
Carlini, N., Ippolito, D., Jagielski, M., Lee, K., Tramer, F., and Zhang, C.
\newblock Quantifying memorization across neural language models.
\newblock \emph{arXiv preprint arXiv:2202.07646}, 2022{\natexlab{b}}.

\bibitem[Chen et~al.(2023)Chen, Chen, Goldstein, Huang, and
  Zhou]{chen2023instructzero}
Chen, L., Chen, J., Goldstein, T., Huang, H., and Zhou, T.
\newblock Instructzero: Efficient instruction optimization for black-box large
  language models.
\newblock \emph{arXiv preprint arXiv:2306.03082}, 2023.

\bibitem[Chen et~al.(2021)Chen, Hsieh, and Gong]{chen2021vision}
Chen, X., Hsieh, C.-J., and Gong, B.
\newblock When vision transformers outperform resnets without pre-training or
  strong data augmentations.
\newblock \emph{arXiv preprint arXiv:2106.01548}, 2021.

\bibitem[Da et~al.(2023{\natexlab{a}})Da, Gao, Mei, and Wei]{da2023llm}
Da, L., Gao, M., Mei, H., and Wei, H.
\newblock Llm powered sim-to-real transfer for traffic signal control.
\newblock \emph{arXiv preprint arXiv:2308.14284}, 2023{\natexlab{a}}.

\bibitem[Da et~al.(2023{\natexlab{b}})Da, Liou, Chen, Zhou, Luo, Yang, and
  Wei]{da2023open}
Da, L., Liou, K., Chen, T., Zhou, X., Luo, X., Yang, Y., and Wei, H.
\newblock Open-ti: Open traffic intelligence with augmented language model.
\newblock \emph{arXiv preprint arXiv:2401.00211}, 2023{\natexlab{b}}.

\bibitem[Devlin et~al.(2018)Devlin, Chang, Lee, and Toutanova]{devlin2018bert}
Devlin, J., Chang, M.-W., Lee, K., and Toutanova, K.
\newblock Bert: Pre-training of deep bidirectional transformers for language
  understanding.
\newblock \emph{arXiv preprint arXiv:1810.04805}, 2018.

\bibitem[Dong et~al.(2019)Dong, Roth, and Su]{dong2019gaussian}
Dong, J., Roth, A., and Su, W.~J.
\newblock Gaussian differential privacy.
\newblock \emph{arXiv preprint arXiv:1905.02383}, 2019.

\bibitem[Du et~al.(2021)Du, Yan, Feng, Zhou, Zhen, Goh, and
  Tan]{du2021efficient}
Du, J., Yan, H., Feng, J., Zhou, J.~T., Zhen, L., Goh, R. S.~M., and Tan, V.~Y.
\newblock Efficient sharpness-aware minimization for improved training of
  neural networks.
\newblock \emph{arXiv preprint arXiv:2110.03141}, 2021.

\bibitem[Du et~al.(2023)Du, Yue, Chow, Wang, Huang, and Sun]{du2023dp}
Du, M., Yue, X., Chow, S.~S., Wang, T., Huang, C., and Sun, H.
\newblock Dp-forward: Fine-tuning and inference on language models with
  differential privacy in forward pass.
\newblock In \emph{Proceedings of the 2023 ACM SIGSAC Conference on Computer
  and Communications Security}, pp.\  2665--2679, 2023.

\bibitem[Duan et~al.(2023)Duan, Dziedzic, Papernot, and
  Boenisch]{duan2023flocks}
Duan, H., Dziedzic, A., Papernot, N., and Boenisch, F.
\newblock Flocks of stochastic parrots: Differentially private prompt learning
  for large language models.
\newblock \emph{arXiv preprint arXiv:2305.15594}, 2023.

\bibitem[Dupuy et~al.(2022)Dupuy, Arava, Gupta, and
  Rumshisky]{dupuy2022efficient}
Dupuy, C., Arava, R., Gupta, R., and Rumshisky, A.
\newblock An efficient dp-sgd mechanism for large scale nlu models.
\newblock In \emph{ICASSP 2022-2022 IEEE International Conference on Acoustics,
  Speech and Signal Processing (ICASSP)}, pp.\  4118--4122. IEEE, 2022.

\bibitem[Dwork(2006)]{dwork2006differential}
Dwork, C.
\newblock Differential privacy.
\newblock In \emph{International colloquium on automata, languages, and
  programming}, pp.\  1--12. Springer, 2006.

\bibitem[Dwork et~al.(2014)Dwork, Roth, et~al.]{dwork2014algorithmic}
Dwork, C., Roth, A., et~al.
\newblock The algorithmic foundations of differential privacy.
\newblock \emph{Foundations and Trends{\textregistered} in Theoretical Computer
  Science}, 9\penalty0 (3--4):\penalty0 211--407, 2014.

\bibitem[Elmahdy \& Salem(2023)Elmahdy and Salem]{elmahdy2023deconstructing}
Elmahdy, A. and Salem, A.
\newblock Deconstructing classifiers: Towards a data reconstruction attack
  against text classification models.
\newblock \emph{arXiv preprint arXiv:2306.13789}, 2023.

\bibitem[Gou et~al.(2021)Gou, Yu, Maybank, and Tao]{gou2021knowledge}
Gou, J., Yu, B., Maybank, S.~J., and Tao, D.
\newblock Knowledge distillation: A survey.
\newblock \emph{International Journal of Computer Vision}, 129:\penalty0
  1789--1819, 2021.

\bibitem[Huang et~al.(2022)Huang, Shao, and Chang]{huang2022large}
Huang, J., Shao, H., and Chang, K. C.-C.
\newblock Are large pre-trained language models leaking your personal
  information?
\newblock \emph{arXiv preprint arXiv:2205.12628}, 2022.

\bibitem[Ippolito et~al.(2022)Ippolito, Tram{\`e}r, Nasr, Zhang, Jagielski,
  Lee, Choquette-Choo, and Carlini]{ippolito2022preventing}
Ippolito, D., Tram{\`e}r, F., Nasr, M., Zhang, C., Jagielski, M., Lee, K.,
  Choquette-Choo, C.~A., and Carlini, N.
\newblock Preventing verbatim memorization in language models gives a false
  sense of privacy.
\newblock \emph{arXiv preprint arXiv:2210.17546}, 2022.

\bibitem[Lester et~al.(2021)Lester, Al-Rfou, and Constant]{lester2021power}
Lester, B., Al-Rfou, R., and Constant, N.
\newblock The power of scale for parameter-efficient prompt tuning.
\newblock \emph{arXiv preprint arXiv:2104.08691}, 2021.

\bibitem[Li et~al.(2023{\natexlab{a}})Li, Xu, and Song]{li2023sentence}
Li, H., Xu, M., and Song, Y.
\newblock Sentence embedding leaks more information than you expect: Generative
  embedding inversion attack to recover the whole sentence.
\newblock \emph{arXiv preprint arXiv:2305.03010}, 2023{\natexlab{a}}.

\bibitem[Li et~al.(2021)Li, Tramer, Liang, and Hashimoto]{li2021large}
Li, X., Tramer, F., Liang, P., and Hashimoto, T.
\newblock Large language models can be strong differentially private learners.
\newblock \emph{arXiv preprint arXiv:2110.05679}, 2021.

\bibitem[Li \& Liang(2021)Li and Liang]{li2021prefix}
Li, X.~L. and Liang, P.
\newblock Prefix-tuning: Optimizing continuous prompts for generation.
\newblock \emph{arXiv preprint arXiv:2101.00190}, 2021.

\bibitem[Li et~al.(2023{\natexlab{b}})Li, Tan, and Liu]{li2023privacy}
Li, Y., Tan, Z., and Liu, Y.
\newblock Privacy-preserving prompt tuning for large language model services.
\newblock \emph{arXiv preprint arXiv:2305.06212}, 2023{\natexlab{b}}.

\bibitem[Liu et~al.(2021)Liu, Ji, Fu, Tam, Du, Yang, and Tang]{liu2021p}
Liu, X., Ji, K., Fu, Y., Tam, W.~L., Du, Z., Yang, Z., and Tang, J.
\newblock P-tuning v2: Prompt tuning can be comparable to fine-tuning
  universally across scales and tasks.
\newblock \emph{arXiv preprint arXiv:2110.07602}, 2021.

\bibitem[Liu et~al.(2023)Liu, Zheng, Du, Ding, Qian, Yang, and
  Tang]{liu2023gpt}
Liu, X., Zheng, Y., Du, Z., Ding, M., Qian, Y., Yang, Z., and Tang, J.
\newblock Gpt understands, too.
\newblock \emph{AI Open}, 2023.

\bibitem[Liu et~al.(2019)Liu, Ott, Goyal, Du, Joshi, Chen, Levy, Lewis,
  Zettlemoyer, and Stoyanov]{liu2019roberta}
Liu, Y., Ott, M., Goyal, N., Du, J., Joshi, M., Chen, D., Levy, O., Lewis, M.,
  Zettlemoyer, L., and Stoyanov, V.
\newblock Roberta: A robustly optimized bert pretraining approach.
\newblock \emph{arXiv preprint arXiv:1907.11692}, 2019.

\bibitem[Luo et~al.(2022)Luo, Lau, Li, and Si]{luo2022critical}
Luo, B., Lau, R.~Y., Li, C., and Si, Y.-W.
\newblock A critical review of state-of-the-art chatbot designs and
  applications.
\newblock \emph{Wiley Interdisciplinary Reviews: Data Mining and Knowledge
  Discovery}, 12\penalty0 (1):\penalty0 e1434, 2022.

\bibitem[Lyu et~al.(2020)Lyu, He, and Li]{lyu2020differentially}
Lyu, L., He, X., and Li, Y.
\newblock Differentially private representation for nlp: Formal guarantee and
  an empirical study on privacy and fairness.
\newblock \emph{arXiv preprint arXiv:2010.01285}, 2020.

\bibitem[Malladi et~al.(2023)Malladi, Gao, Nichani, Damian, Lee, Chen, and
  Arora]{malladi2023fine}
Malladi, S., Gao, T., Nichani, E., Damian, A., Lee, J.~D., Chen, D., and Arora,
  S.
\newblock Fine-tuning language models with just forward passes.
\newblock \emph{arXiv preprint arXiv:2305.17333}, 2023.

\bibitem[Mialon et~al.(2023)Mialon, Dess{\`\i}, Lomeli, Nalmpantis, Pasunuru,
  Raileanu, Rozi{\`e}re, Schick, Dwivedi-Yu, Celikyilmaz,
  et~al.]{mialon2023augmented}
Mialon, G., Dess{\`\i}, R., Lomeli, M., Nalmpantis, C., Pasunuru, R., Raileanu,
  R., Rozi{\`e}re, B., Schick, T., Dwivedi-Yu, J., Celikyilmaz, A., et~al.
\newblock Augmented language models: a survey.
\newblock \emph{arXiv preprint arXiv:2302.07842}, 2023.

\bibitem[Mireshghallah et~al.(2022)Mireshghallah, Uniyal, Wang, Evans, and
  Berg-Kirkpatrick]{mireshghallah2022memorization}
Mireshghallah, F., Uniyal, A., Wang, T., Evans, D., and Berg-Kirkpatrick, T.
\newblock Memorization in nlp fine-tuning methods.
\newblock \emph{arXiv preprint arXiv:2205.12506}, 2022.

\bibitem[Mironov(2017)]{mironov2017renyi}
Mironov, I.
\newblock R{\'e}nyi differential privacy.
\newblock In \emph{2017 IEEE 30th computer security foundations symposium
  (CSF)}, pp.\  263--275. IEEE, 2017.

\bibitem[Nan et~al.(2020)Nan, Radev, Zhang, Rau, Sivaprasad, Hsieh, Tang, Vyas,
  Verma, Krishna, et~al.]{nan2020dart}
Nan, L., Radev, D., Zhang, R., Rau, A., Sivaprasad, A., Hsieh, C., Tang, X.,
  Vyas, A., Verma, N., Krishna, P., et~al.
\newblock Dart: Open-domain structured data record to text generation.
\newblock \emph{arXiv preprint arXiv:2007.02871}, 2020.

\bibitem[Novikova et~al.(2017)Novikova, Du{\v{s}}ek, and
  Rieser]{novikova2017e2e}
Novikova, J., Du{\v{s}}ek, O., and Rieser, V.
\newblock The e2e dataset: New challenges for end-to-end generation.
\newblock \emph{arXiv preprint arXiv:1706.09254}, 2017.

\bibitem[OpenAI(2023)]{OpenAI_GPT4_2023}
OpenAI.
\newblock Gpt-4 technical report.
\newblock \emph{ArXiv}, abs/2303.08774, 2023.
\newblock URL \url{https://arxiv.org/abs/2303.08774}.

\bibitem[Pan et~al.(2020)Pan, Zhang, Ji, and Yang]{pan2020privacy}
Pan, X., Zhang, M., Ji, S., and Yang, M.
\newblock Privacy risks of general-purpose language models.
\newblock In \emph{2020 IEEE Symposium on Security and Privacy (SP)}, pp.\
  1314--1331. IEEE, 2020.

\bibitem[Parikh et~al.(2022)Parikh, Dupuy, and Gupta]{parikh2022canary}
Parikh, R., Dupuy, C., and Gupta, R.
\newblock Canary extraction in natural language understanding models.
\newblock \emph{arXiv preprint arXiv:2203.13920}, 2022.

\bibitem[Radford et~al.(2019)Radford, Wu, Child, Luan, Amodei, Sutskever,
  et~al.]{radford2019language}
Radford, A., Wu, J., Child, R., Luan, D., Amodei, D., Sutskever, I., et~al.
\newblock Language models are unsupervised multitask learners.
\newblock \emph{OpenAI blog}, 1\penalty0 (8):\penalty0 9, 2019.

\bibitem[Scao \& Rush(2021)Scao and Rush]{scao2021many}
Scao, T.~L. and Rush, A.~M.
\newblock How many data points is a prompt worth?
\newblock \emph{arXiv preprint arXiv:2103.08493}, 2021.

\bibitem[Shin et~al.(2020)Shin, Razeghi, Logan~IV, Wallace, and
  Singh]{shin2020autoprompt}
Shin, T., Razeghi, Y., Logan~IV, R.~L., Wallace, E., and Singh, S.
\newblock Autoprompt: Eliciting knowledge from language models with
  automatically generated prompts.
\newblock \emph{arXiv preprint arXiv:2010.15980}, 2020.

\bibitem[Shokri et~al.(2017)Shokri, Stronati, Song, and
  Shmatikov]{shokri2017membership}
Shokri, R., Stronati, M., Song, C., and Shmatikov, V.
\newblock Membership inference attacks against machine learning models.
\newblock In \emph{2017 IEEE symposium on security and privacy (SP)}, pp.\
  3--18. IEEE, 2017.

\bibitem[Socher et~al.(2013)Socher, Perelygin, Wu, Chuang, Manning, Ng, and
  Potts]{socher2013recursive}
Socher, R., Perelygin, A., Wu, J., Chuang, J., Manning, C.~D., Ng, A.~Y., and
  Potts, C.
\newblock Recursive deep models for semantic compositionality over a sentiment
  treebank.
\newblock In \emph{Proceedings of the 2013 conference on empirical methods in
  natural language processing}, pp.\  1631--1642, 2013.

\bibitem[Song \& Raghunathan(2020)Song and Raghunathan]{song2020information}
Song, C. and Raghunathan, A.
\newblock Information leakage in embedding models.
\newblock In \emph{Proceedings of the 2020 ACM SIGSAC conference on computer
  and communications security}, pp.\  377--390, 2020.

\bibitem[Spall(1992)]{spall1992multivariate}
Spall, J.~C.
\newblock Multivariate stochastic approximation using a simultaneous
  perturbation gradient approximation.
\newblock \emph{IEEE transactions on automatic control}, 37\penalty0
  (3):\penalty0 332--341, 1992.

\bibitem[Sun et~al.(2022)Sun, Shao, Qian, Huang, and Qiu]{sun2022black}
Sun, T., Shao, Y., Qian, H., Huang, X., and Qiu, X.
\newblock Black-box tuning for language-model-as-a-service.
\newblock In \emph{International Conference on Machine Learning}, pp.\
  20841--20855. PMLR, 2022.

\bibitem[Tang et~al.(2024)Tang, Panda, Nasr, Mahloujifar, and
  Mittal]{tang2024private}
Tang, X., Panda, A., Nasr, M., Mahloujifar, S., and Mittal, P.
\newblock Private fine-tuning of large language models with zeroth-order
  optimization.
\newblock \emph{arXiv preprint arXiv:2401.04343}, 2024.

\bibitem[Touvron et~al.(2023)Touvron, Martin, Stone, Albert, Almahairi, Babaei,
  Bashlykov, Batra, Bhargava, Bhosale, et~al.]{touvron2023llama}
Touvron, H., Martin, L., Stone, K., Albert, P., Almahairi, A., Babaei, Y.,
  Bashlykov, N., Batra, S., Bhargava, P., Bhosale, S., et~al.
\newblock Llama 2: Open foundation and fine-tuned chat models.
\newblock \emph{arXiv preprint arXiv:2307.09288}, 2023.

\bibitem[Voorhees et~al.(1999)]{voorhees1999trec}
Voorhees, E.~M. et~al.
\newblock The trec-8 question answering track report.
\newblock In \emph{Trec}, volume~99, pp.\  77--82, 1999.

\bibitem[Wang et~al.(2018)Wang, Singh, Michael, Hill, Levy, and
  Bowman]{wang2018glue}
Wang, A., Singh, A., Michael, J., Hill, F., Levy, O., and Bowman, S.~R.
\newblock Glue: A multi-task benchmark and analysis platform for natural
  language understanding.
\newblock \emph{arXiv preprint arXiv:1804.07461}, 2018.

\bibitem[Wang et~al.(2023)Wang, Chen, Pei, Xie, Kang, Zhang, Xu, Xiong, Dutta,
  Schaeffer, et~al.]{wang2023decodingtrust}
Wang, B., Chen, W., Pei, H., Xie, C., Kang, M., Zhang, C., Xu, C., Xiong, Z.,
  Dutta, R., Schaeffer, R., et~al.
\newblock Decodingtrust: A comprehensive assessment of trustworthiness in gpt
  models.
\newblock \emph{arXiv preprint arXiv:2306.11698}, 2023.

\bibitem[Williams et~al.(2017)Williams, Nangia, and Bowman]{williams2017broad}
Williams, A., Nangia, N., and Bowman, S.~R.
\newblock A broad-coverage challenge corpus for sentence understanding through
  inference.
\newblock \emph{arXiv preprint arXiv:1704.05426}, 2017.

\bibitem[Wu et~al.(2020)Wu, Xia, and Wang]{wu2020adversarial}
Wu, D., Xia, S.-T., and Wang, Y.
\newblock Adversarial weight perturbation helps robust generalization.
\newblock \emph{Advances in Neural Information Processing Systems},
  33:\penalty0 2958--2969, 2020.

\bibitem[Yang et~al.(2022{\natexlab{a}})Yang, Tian, Peng, and
  Klein]{yang2022re3}
Yang, K., Tian, Y., Peng, N., and Klein, D.
\newblock Re3: Generating longer stories with recursive reprompting and
  revision.
\newblock \emph{arXiv preprint arXiv:2210.06774}, 2022{\natexlab{a}}.

\bibitem[Yang et~al.(2022{\natexlab{b}})Yang, Chen, PourNejatian, Shin, Smith,
  Parisien, Compas, Martin, Flores, Zhang, et~al.]{yang2022gatortron}
Yang, X., Chen, A., PourNejatian, N., Shin, H.~C., Smith, K.~E., Parisien, C.,
  Compas, C., Martin, C., Flores, M.~G., Zhang, Y., et~al.
\newblock Gatortron: A large clinical language model to unlock patient
  information from unstructured electronic health records.
\newblock \emph{arXiv preprint arXiv:2203.03540}, 2022{\natexlab{b}}.

\bibitem[Yeom et~al.(2018)Yeom, Giacomelli, Fredrikson, and
  Jha]{yeom2018privacy}
Yeom, S., Giacomelli, I., Fredrikson, M., and Jha, S.
\newblock Privacy risk in machine learning: Analyzing the connection to
  overfitting.
\newblock In \emph{2018 IEEE 31st computer security foundations symposium
  (CSF)}, pp.\  268--282. IEEE, 2018.

\bibitem[Yu et~al.(2021{\natexlab{a}})Yu, Naik, Backurs, Gopi, Inan, Kamath,
  Kulkarni, Lee, Manoel, Wutschitz, et~al.]{yu2021differentially}
Yu, D., Naik, S., Backurs, A., Gopi, S., Inan, H.~A., Kamath, G., Kulkarni, J.,
  Lee, Y.~T., Manoel, A., Wutschitz, L., et~al.
\newblock Differentially private fine-tuning of language models.
\newblock \emph{arXiv preprint arXiv:2110.06500}, 2021{\natexlab{a}}.

\bibitem[Yu et~al.(2021{\natexlab{b}})Yu, Zhang, Chen, Yin, and
  Liu]{yu2021large}
Yu, D., Zhang, H., Chen, W., Yin, J., and Liu, T.-Y.
\newblock Large scale private learning via low-rank reparametrization.
\newblock In \emph{International Conference on Machine Learning}, pp.\
  12208--12218. PMLR, 2021{\natexlab{b}}.

\bibitem[Zhang et~al.(2023)Zhang, Thekumparampil, Oh, and He]{zhang2023dpzero}
Zhang, L., Thekumparampil, K.~K., Oh, S., and He, N.
\newblock Dpzero: Dimension-independent and differentially private zeroth-order
  optimization.
\newblock \emph{arXiv preprint arXiv:2310.09639}, 2023.

\bibitem[Zhang et~al.(2022{\natexlab{a}})Zhang, Hidano, and
  Koushanfar]{zhang2022text}
Zhang, R., Hidano, S., and Koushanfar, F.
\newblock Text revealer: Private text reconstruction via model inversion
  attacks against transformers.
\newblock \emph{arXiv preprint arXiv:2209.10505}, 2022{\natexlab{a}}.

\bibitem[Zhang et~al.(2022{\natexlab{b}})Zhang, Yao, Jia, Yi, Hong, Chang, and
  Liu]{zhang2022robustify}
Zhang, Y., Yao, Y., Jia, J., Yi, J., Hong, M., Chang, S., and Liu, S.
\newblock How to robustify black-box ml models? a zeroth-order optimization
  perspective.
\newblock \emph{arXiv preprint arXiv:2203.14195}, 2022{\natexlab{b}}.

\bibitem[Zhou et~al.(2023)Zhou, Jiang, Cui, Wang, Xiao, Hou, Cotterell, and
  Sachan]{zhou2023recurrentgpt}
Zhou, W., Jiang, Y.~E., Cui, P., Wang, T., Xiao, Z., Hou, Y., Cotterell, R.,
  and Sachan, M.
\newblock Recurrentgpt: Interactive generation of (arbitrarily) long text.
\newblock \emph{arXiv preprint arXiv:2305.13304}, 2023.

\end{thebibliography}
\bibliographystyle{icml2023}

\newpage
\appendix
\onecolumn

\section{Experiment Setups}

\subsection{Hyperparameters} 
\label{app:hype}
Different tasks and methods require different parameters. For example, full fine-tuning requires a much smaller learning rate while prefix tuning needs a much larger learning rate. Besides, tasks like table-to-text generation require a small learning with a large training epoch. The only fixed hyperparameter is the batch size. 
We set the batch size to 1024 for all settings with gradient accumulation. Detailed hyperparameters for MNLI and E2E can be found in \cref{tab:main_hyperparameter}. For \ours, we set the regularization weight $\lambda$ in \cref{eq:final} to 0.01 for all experiments. For the DP algorithms, we follow the common practice to set the privacy budget as $\epsilon = [3,8]$ and $\delta = 1e-5$ for all settings, and we account for privacy through Renyi differential privacy~\cite{mironov2017renyi}, the code will be released on author's website.


\begin{table}[h!]
\centering
\begin{tabular}{|l|c|c|}
\hline
Methods & Learning Rate  & Training Epoch  \\
\hline
\multicolumn{3}{|c|}{Non private-MNLI} \\
\hline
Full Fine-tuning & 5e-5  & 5 \\
Prefix Tuning & 0.01 & 20 \\
\ours & 0.01 & 20 \\
\hline
\multicolumn{3}{|c|}{DP setting-MNLI} \\
\hline
Full Fine-tuning & 5e-4  & 5 \\
Prefix Tuning & 0.01 & 20 \\
\ours & 0.01 & 20 \\
\hline
\multicolumn{3}{|c|}{Non private-E2E} \\
\hline
Full Fine-tuning & 2e-3  & 15 \\
Prefix Tuning & 5e-4 & 30 \\
\ours & 5e-4 & 30 \\
\hline
\multicolumn{3}{|c|}{DP setting-E2E} \\
\hline
Full Fine-tuning & 2e-3  & 15 \\
Prefix Tuning & 5e-4 & 100 \\
\ours & 5e-4 & 100 \\
\hline
\end{tabular}
\caption{Detailed hyperparameters for DP training and normal training on MNLI and E2E.}

\label{tab:main_hyperparameter}
\end{table}

\subsection{Settings for Membership Inference Attack}
\label{app:mia}
We evaluate the privacy risks empirically by membership inference attack (MIA) using Likelihood Ratio test (LiRA) (Mireshghallah et al., 2022). For SST-2, because of the distributional bias between the training and test sets, we filter the training set to include samples with more than 20 tokens, in which case only 15 test samples are eliminated. The data filtering can avoid undesired high MIA accuracy due to the lack of short samples in test sets. Then we compute the loss for all samples in $\hat{\mathcal{D}}$ and rank every sample by its loss. We label all the samples with $1\%$ lowest loss as training data and compute the success rate of MIA only on samples with $1\%$ lowest loss. Note that a model that preserves more privacy indicates that the success rate of MIA is closer to $50\%$ because if attackers get an MIA success rate below $50\%$, they could use reverse results to implement attacks.
The results are reported in \cref{tab:MIA} under text classification datasets with both white-box and black-box settings.  


\end{document}